\documentclass[11pt]{article}

\usepackage{arxiv}

\usepackage[utf8]{inputenc} 
\usepackage[T1]{fontenc}    
\usepackage{hyperref}       
\usepackage{url}            
\usepackage{booktabs}       
\usepackage{array}
\usepackage{longtable}
\usepackage{graphicx}
\usepackage{multirow}
\usepackage{amsfonts}       
\usepackage{microtype}      

\title{AgentBay: A Hybrid Interaction Sandbox for Seamless Human-AI Intervention in Agentic Systems}

\date{}

\renewcommand{\headeright}{}
\renewcommand{\undertitle}{}
\renewcommand{\shorttitle}{AgentBay: A Hybrid Interaction Sandbox}

\renewenvironment{abstract}
{
  \centerline
  {\large \bfseries \scshape Abstract}
  \begin{list}{}{%
    \setlength{\leftmargin}{2em}      
    \setlength{\rightmargin}{2em}     
    \setlength{\listparindent}{0pt}   
    \setlength{\itemindent}{0pt}      
    \setlength{\parsep}{0pt}          
    \setlength{\topsep}{0pt}          
  }
  \item\relax
}
{
  \end{list}
}

\makeatletter
\renewcommand{\@maketitle}{%
  \vbox{%
    \hsize\textwidth
    \linewidth\hsize
    \vskip 0.1in
    \@toptitlebar
    \centering
    {\LARGE\bfseries \@title\par}  
    \@bottomtitlebar
    \textsc{\undertitle}\\
    \vskip 0.1in
    \def\And{%
      \end{tabular}\hfil\linebreak[0]\hfil%
      \begin{tabular}[t]{c}\bf\rule{\z@}{24\p@}\ignorespaces%
    }
    \def\AND{%
      \end{tabular}\hfil\linebreak[4]\hfil%
      \begin{tabular}[t]{c}\bf\rule{\z@}{24\p@}\ignorespaces%
    }
    \begin{tabular}[t]{c}\bf\rule{\z@}{24\p@}\@author\end{tabular}%
  \vskip 0.4in \@minus 0.1in \center{\@date}   \vskip 0.2in
  }
}
\makeatother

\hypersetup{
pdftitle={AgentBay: A Hybrid Interaction Sandbox for Seamless Human-AI Intervention in Agentic Systems},
pdfsubject={cs.AI, cs.SE},
pdfauthor={},
pdfkeywords={AI Agents, Human-in-the-Loop, Sandbox, Adaptive Streaming Protocol},
}

\author{
  Yun Piao, Hongbo Min, Hang Su, Leilei Zhang, Lei Wang, Yue Yin, Xiao Wu, Zhejing Xu, \\
  \textbf{Liwei Qu, Hang Li, Xinxin Zeng, Wei Tian, Fei Yu, Xiaowei Li, Jiayi Jiang, Tongxu Liu,} \\
  \textbf{Hao Tian, Yufei Que, Xiaobing Tu, Bing Suo, Yuebing Li, Xiangting Chen, Zeen Zhao, Jiaming Tang,} \\
  \textbf{Wei Huang, Xuguang Li, Jing Zhao, Jin Li, Jie Shen, Jinkui Ren, Xiantao Zhang} \\
  \\
  Alibaba Cloud Computing
}

\AtBeginDocument{%
  \setlength{\leftmargini}{1em}
  \setlength{\labelwidth}{1em}
  \setlength{\labelsep}{0.5em}
  \setlength{\itemindent}{0pt}
  \setlength{\leftmarginii}{2.5em}
  \setlength{\leftmarginiii}{2.5em}%
  \setlength{\leftmarginiv}{2.5em}%
  \setlength{\leftmarginv}{2.5em}%
  \setlength{\leftmarginvi}{2.5em}%
  \def\@listi{\leftmargin\leftmargini
              \labelwidth\labelwidth
              \topsep 4\p@ \@plus 1\p@ \@minus 2\p@
              \parsep 2\p@ \@plus 1\p@ \@minus 0.5\p@
              \itemsep \parsep}%
}

\begin{document}

\maketitle

\begin{abstract}
The rapid advancement of Large Language Models (LLMs) is catalyzing a shift towards autonomous AI Agents capable of executing complex, multi-step tasks. However, these agents remain brittle when faced with real-world exceptions, making Human-in-the-Loop (HITL) supervision essential for mission-critical applications. In this paper, we present AgentBay, a novel sandbox service designed from the ground up for hybrid interaction. AgentBay provides secure, isolated execution environments spanning Windows, Linux, Android, Web Browsers, and Code interpreters. Its core contribution is a unified session accessible via a hybrid control interface: An AI agent can interact programmatically via mainstream interfaces (MCP, Open Source SDK), while a human operator can, at any moment, seamlessly take over full manual control.

This seamless intervention is enabled by Adaptive Streaming Protocol (ASP). Unlike traditional VNC/RDP, ASP is specifically engineered for this hybrid use case, delivering an ultra-low-latency, smoother user experience that remains resilient even in weak network environments. It achieves this by dynamically blending command-based and video-based streaming, adapting its encoding strategy based on network conditions and the current controller (AI or human).

Our evaluation demonstrates strong results in security, performance, and task completion rates. In a benchmark of complex tasks, the AgentBay (Agent + Human) model achieved more than 48\% success rate improvement. Furthermore, our ASP protocol reduces bandwidth consumption by up to 50\% compared to standard RDP, and in end-to-end latency with around 5\% reduction, especially under poor network conditions. We posit that AgentBay provides a foundational primitive for building the next generation of reliable, human-supervised autonomous systems.
\end{abstract}

\vspace{0.5em plus 0.2em minus 0.1em}

\section{Introduction}

The proliferation of Large Language Models (LLMs) has given rise to a new paradigm of autonomous agents~\cite{placeholder-llm-agent-survey}. Systems like Auto-GPT~\cite{placeholder-autogpt} and frameworks like LangChain~\cite{placeholder-langchain} empower agents to reason, plan, and execute tasks across digital environments. However, their deployment is hindered by two key challenges: brittleness and the need for secure human-in-the-loop (HITL) collaboration. Brittleness occurs when agents fail at unforeseen exceptions---modal pop-ups or CAPTCHAs~\cite{placeholder-captcha}. The collaboration need arises when agents must securely handle private data. For example, an agent might autonomously navigate a website, but upon reaching the login page, it must pause and request human intervention to securely enter credentials (username and password), as the agent itself is not provisioned with this sensitive information. In both scenarios---unexpected failure or planned intervention---a seamless, low-friction handoff to a human operator is critical.

This necessity for supervision highlights the critical importance of sandboxed environments that also support fluid human intervention. Leading agent sandboxes (e.g., E2B~\cite{placeholder-e2b}, Daytona~\cite{placeholder-daytona}) recognize this need, and they typically address it by integrating general-purpose remote interaction protocols like VNC~\cite{placeholder-vnc} or RDP~\cite{placeholder-rdp} as their human intervention mechanism. While this approach provides a vital function, these protocols were not specifically designed for the rapid, low-friction handoff required in hybrid AI systems. They can introduce noticeable latency and may be less resilient under variable network conditions, which can diminish the fluidity of the human-agent collaboration.

To address this specific challenge, we present a hybrid interaction sandbox infrastructure designed from the ground up to optimize this interaction (based on the production service AgentBay). AgentBay provides a single, isolated execution sandbox that can be controlled simultaneously via a programmatic API and open source SDK for AI Agent, or a high-performance graphical streaming interface for humans.

The core of our system is the Adaptive Streaming Protocol (ASP). When a human takes over, it instantly switches to an ultra-low-latency, smooth, and resilient graphical stream explicitly optimized for interaction fluency, even on variable networks.

We make the following contributions:

\begin{itemize}
  \item \textbf{Hybrid Interaction Architecture:} We present the design of a hybrid interaction sandbox system supporting diverse OS, mobile, browser, and code, that are secure, isolated execution environments.
  \item \textbf{Adaptive Streaming Protocol (ASP):} We detail our novel streaming protocol that enables human control, specifically engineered to improve key metrics over traditional protocols in this use case.
  \item \textbf{Comprehensive Evaluation (with hypothetical data):} We demonstrate that:
  \begin{itemize}
    \item The hybrid interaction (Agent + HITL) model drastically improves task success rates on complex benchmarks.
    \item ASP significantly outperforms traditional protocols (e.g., RDP) in bandwidth efficiency---achieving up to 50\% reduction---and in interaction latency, with around 5\% reduction, especially under poor network conditions.
  \end{itemize}
\end{itemize}

We have released an open-source SDK~\cite{placeholder-agentbay-sdk} to interface with AgentBay on GitHub, encouraging community adoption and further research.

\section{Related Work}

The emergence of autonomous AI agents has spurred research across multiple interconnected domains, including agent architectures, secure execution environments, human-in-the-loop (HITL) interaction paradigms, and remote display technologies. While significant progress has been made in each area, a critical gap remains: the lack of a unified infrastructure that simultaneously supports programmatic agent control, secure multi-platform execution, and seamless, real-time human intervention. Existing solutions tend to optimize for one or two of these dimensions while neglecting the others---particularly the fluid handoff between autonomous and manual control. In what follows, we situate AgentBay within this landscape by reviewing relevant work along four axes.

The figure below illustrates the interaction among the core components of the system. The user provides a task to an agent that reasons using the ReAct paradigm. The agent leverages a language model to deliberate and formulate concrete tool-invocation actions. These tools are securely executed within remote sandboxed environments, and their execution results are returned to the agent. During execution, visual states from the sandbox can be streamed in real time to both the user and the agent, serving as perceptual inputs to inform subsequent decision-making. If necessary, the user can intervene by manually triggering actions, thereby enabling human-in-the-loop control.

\begin{figure}[htbp]
  \centering
  \includegraphics[width=0.8\linewidth]{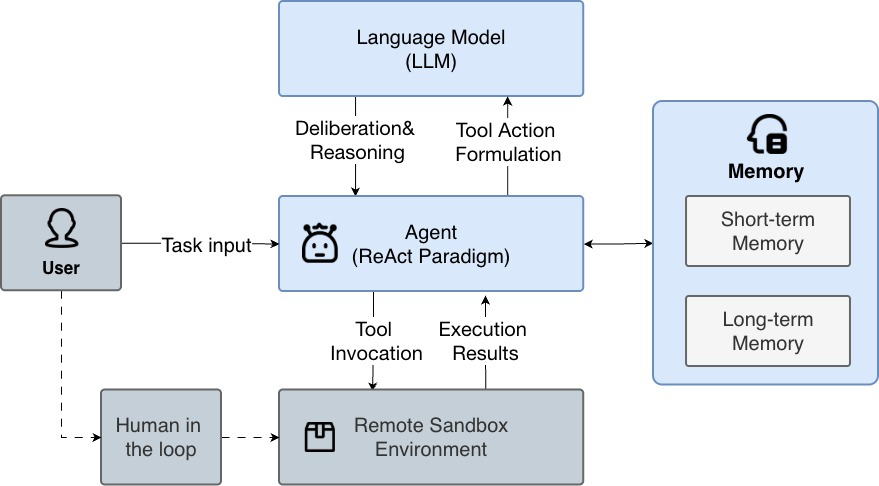}
  \caption{AI Agent with Sandboxes.}
  \label{fig:ai-agent-sandboxes}
\end{figure}

\noindent\textbf{Autonomous Agent Frameworks:} The agent ecosystem has evolved rapidly. Frameworks like LangChain~\cite{placeholder-langchain} and LlamaIndex~\cite{placeholder-llamaindex} provide modular components. Foundational agent architectures like ReAct and Auto-GPT combine reasoning and acting. Our work does not propose a new agent architecture; rather, it provides the robust execution environment (or ``actuator'') these agents require to operate reliably in the real world.

\noindent\textbf{Agent Execution Environments:} Recognizing the need for safe execution, several sandboxes have emerged. Systems like E2B provide secure, API-driven Linux sandboxes, primarily for code execution. Daytona focuses on standardizing development environments, which agents can use. These systems excel at programmatic control but are not designed for, and do not provide, seamless human graphical intervention.

\noindent\textbf{Human-in-the-Loop (HITL) Systems:} HITL is a well-established field, particularly in data labeling and active learning. However, in the context of autonomous agents, HITL often means simple ``yes/no'' confirmation prompts at a planning stage. Our work enables a much deeper, more granular level of intervention: full, real-time, graphical control at any point in the agent's execution loop.

\noindent\textbf{Remote Display Protocols:} The performance of human interaction hinges on the streaming protocol. VNC (RFB protocol) is simple and universal but inefficient, relying on raw framebuffer updates. RDP is highly optimized for Windows GDI commands but struggles with non-GDI content (e.g., video, 3D graphics) and is platform-specific. WebRTC is excellent for low-latency video streaming but is not optimized for static desktop content, often consuming unnecessary bandwidth. None are engineered to support the dynamic, controller-aware adaptation required for seamless hybrid interaction.

Below are some representative use cases for secure agent execution.

\begin{itemize}
  \item \textbf{Autonomous Coding Agent:} LLM-powered coding agents (e.g., GitHub Copilot) can interpret natural language prompts and auto-generate, install dependencies for, and validate code. However, outputs may contain bugs, infinite loops, resource leaks, or even malicious commands like \texttt{os.system('rm -fr /')}. Direct execution risks system stability and data security. Sandboxing prevents local damage by isolating file access, limiting CPU memory usage, and containing crashes. It also enables closed-loop feedback---capturing stdout, stderr, and exceptions (e.g., ``RecursionError'') to guide agent self-correction---and ensures multi-tenant isolation in shared platforms. Tang et al. propose CodeAgent, an autonomous communicative agent designed for code review, and detail its architecture and implementation~\cite{placeholder-codeagent}.

  \item \textbf{Data Science Agent:} Data science agents let non-experts request analyses like ``plot quarterly sales'' in natural language and auto-execute Pandas/Matplotlib scripts. But they handle sensitive data and risk memory overflows or accidental data leaks if run openly. Sandboxes protect privacy by processing data in encrypted, network-isolated environments, enforce resource limits (e.g., 2GB RAM), return visual results via Base64 or secure channels, and use per-task virtual environments to avoid dependency conflicts. Wang et al. present a comprehensive survey on large language model-based data science agents, providing a comprehensive overview of representative frameworks in this emerging area~\cite{placeholder-ds-agent}.

  \item \textbf{Exam/Competition Automatic Assessment Agent:} Modern grading agents use LLMs to dynamically generate test cases and give tailored feedback (e.g., ``fails with duplicate elements''). Yet student submissions may include fork bombs, attempts to read \texttt{/etc/passwd}, or answer theft via environment variables. Sandboxes block dangerous syscalls via seccomp, cgroups, or namespaces; prohibit file writes, networking, and process spawning; ensure consistent runtime environments across users; prevent cheating through strict I/O isolation; and support multiple languages via standardized sandboxed runtimes. Auto-Arena is a fully automated LLM evaluation framework that requires no human intervention, leverages multi-round LLM competitions and collaborative judging, and achieves high alignment with human preferences~\cite{placeholder-auto-arena}.

  \item \textbf{Automated interaction of mobile applications:} LLM-driven mobile agents can perform tasks like ``send `hello' to Zhang San on WeChat'' by parsing UIs and generating taps/swipes. On real devices or non-isolated simulators, this risks accessing contacts, photos, or triggering payments. Sandboxes restrict access to real PII (location, messages), redirect operations to virtual data, isolate financial/social app accounts, comply with GDPR/CCPA by limiting scope, and provide uniform Android/iOS simulation to eliminate device fragmentation in testing. Crucially, Kahlon et al. (2025) observe that even in secure, sandboxed settings, certain tasks cannot be completed without explicit user input~\cite{placeholder-phone-ui}.

  \item \textbf{Game AI training environment:} Games like Minecraft and StarCraft II serve as rich RL training environments, but thousands of trials can cause crashes, memory leaks, or GPU driver issues that destabilize hosts. Shared environments also lead to resource contention. Sandboxed containers isolate game engines (e.g., Unity, Minecraft Server), so crashes don\'t affect the host or other agents. Each RL agent runs in its own sandbox with dedicated resources, fixed random seeds, and enforced quotas (CPU, GPU, FPS), enabling safe, parallel, reproducible training while blocking abuse of in-game scripting interfaces. As highlighted in Hu et al. (2024), game-based environments used for AI training have continuously evolved over time, becoming increasingly sophisticated to better reflect real-world challenges and support advances in artificial intelligence research~\cite{placeholder-game-ai}.
\end{itemize}

\section{Problem Definition}

To systematically address the challenges outlined, we must first deeply analyze the core problems faced by sandboxes in agentic systems.

First and foremost is the \textbf{Problem of Security and Isolation}. AI agents, by design, execute tasks by running code, issuing system commands, and interacting with files. This introduces inherent, significant risks. An autonomous agent could inadvertently execute malicious code, modify or delete critical user files, or access sensitive data, thereby damaging the host environment. Therefore, the foundational challenge is to create a secure, strictly sandboxed environment. This sandbox must enforce robust process, filesystem, and network isolation to ensure that all agent operations are contained, preventing any possibility of ``escaping'' and compromising the user's system.

Second is the \textbf{Problem of Human-AI Handoff and Dual Control}. While agents can automate many steps, they remain ``brittle'' and frequently require human-in-the-loop (HITL) intervention. This need arises in two primary scenarios: planned handoffs (e.g., an agent pausing for a human to securely enter credentials) and unplanned exception handling (e.g., an agent failing at an unforeseen CAPTCHA or UI change, requiring a human to take over). This necessitates a dual-modal control system where an AI agent can operate programmatically via an API, and a human can, at any moment, seamlessly take over via a high-performance graphical stream.

Finally, we face the \textbf{Problem of Environmental Fidelity and Elasticity}. Real-world tasks are not simple and do not run in a single, uniform environment. They are complex and require a heterogeneous ecosystem of high-fidelity, stateful environments, including full desktop operating systems (Windows, Linux), mobile devices (Android), specific browser instances, and code interpreters. Providing and maintaining a persistent fleet of all possible environments is economically and technically infeasible. Therefore, the system must be highly elastic, capable of provisioning these diverse, complex environments on-demand. This ``instant-on'' capability is critical for scalability and for ensuring the agent has the correct, high-fidelity environment required to successfully complete its task.

In summary, the fundamental problem is to design a multi-tenant system that (a) provides verifiable security and isolation for untrusted agent execution, (b) enables seamless, low-friction handoffs between AI and human controllers, and (c) elastically delivers a diverse range of high-fidelity environments to support complex, real-world tasks.

\section{System Architecture}

The AgentBay was architected around several core design principles, aimed at delivering a comprehensive and robust solution for autonomous agent execution.

\begin{itemize}
  \item \textbf{Security and Isolation:} A primary objective is to guarantee a secure execution environment. This is achieved through a multi-tenant architecture that enforces strict process, filesystem, and network isolation for each session. This ``isolated execution'' model ensures that agent operations are securely sandboxed, preventing any impact on the host system or other concurrent sessions.

  \item \textbf{High-Fidelity and Heterogeneous Environments:} A core principle is the provision of authentic, real-world environments, not mere simulations. AgentBay supports a heterogeneous ecosystem with different OS by offering four primary types of mirrored environments: \textbf{Computer Use} (providing a complete desktop OS), \textbf{Mobile Use} (emulating mobile device environments), \textbf{Browser Use} (offering isolated, full-featured browser instances), and \textbf{Code Space} (providing a dedicated development and scripting environment). This commitment to high-fidelity, diverse environments ensures that agents can operate within a stack identical to that of a human user, which is critical for realistic task execution and testing across varied digital scenarios.

  \item \textbf{Hybrid Interaction and Seamless Handoff:} AgentBay is engineered to support a flexible ``Hybrid Interaction'' model. It provides both programmatic (API-driven) access for autonomous tasks and high-performance graphical (streaming) access for human-in-the-loop (HITL) oversight, debugging, or direct intervention. This dual-mode capability is foundational to enabling a ``Seamless Handoff'' between automated control and manual operation within the same persistent session.

  \item \textbf{Performance and Scalability:} The architecture is optimized for high performance and low-latency interaction. This includes rapid provisioning of sandbox environments (on-demand instantiation) and an efficient, high-frame-rate streaming protocol to support responsive graphical interaction. The system is designed to be highly available and horizontally scalable, ensuring reliability for both API-driven tasks and real-time user engagement.
\end{itemize}

\subsection{System Architecture Layers}

The overall technical architecture of AgentBay is systematically structured into four distinct, hierarchical layers, designed to provide a robust and flexible sandbox environment.

\begin{itemize}
  \item \textbf{The Interface Layer:} This layer facilitates interaction and integration with the AgentBay service. It provides open-source Software Development Kits (SDKs) in multiple languages, including TypeScript, Python, and Go, on GitHub to ensure broad developer accessibility. Furthermore, it supports Model Context Protocol (MCP) and integrates Adaptive Streaming Protocol (ASP) for high-performance, low-latency interactions.

  \item \textbf{The Service Layer:} This layer abstracts the core functionalities of the system. It encompasses management services for Sandbox Tasks and Sandbox Tools, orchestrating the execution and resource allocation of virtualized environments. Meanwhile, the streaming service is responsible for rendering and delivering the high-performance, low-latency graphical stream. This service functions as the essential backend for the Adaptive Streaming Protocol (ASP), enabling real-time visual interaction and remote control of the sandboxed environments.

  \item \textbf{The Environment Layer:} This layer provisions the virtualized execution environments. It offers four primary types of mirrored environments to cater to diverse use cases: Browser Use (for web-based scenarios), Code Space (for development and scripting), Computer Use (simulating a standard desktop OS), and Mobile Use (emulating mobile device environments).

  \item \textbf{The Feature Layer:} This foundational layer provides a comprehensive suite of granular capabilities essential for full-featured sandboxing. It is built upon core services including Session management, Network configuration, Context for data persistence, Command execution, File operations, and port mapping, among others, forming a complete and self-contained functional system.
\end{itemize}

\begin{figure}[htbp]
  \centering
  \includegraphics[width=0.9\linewidth]{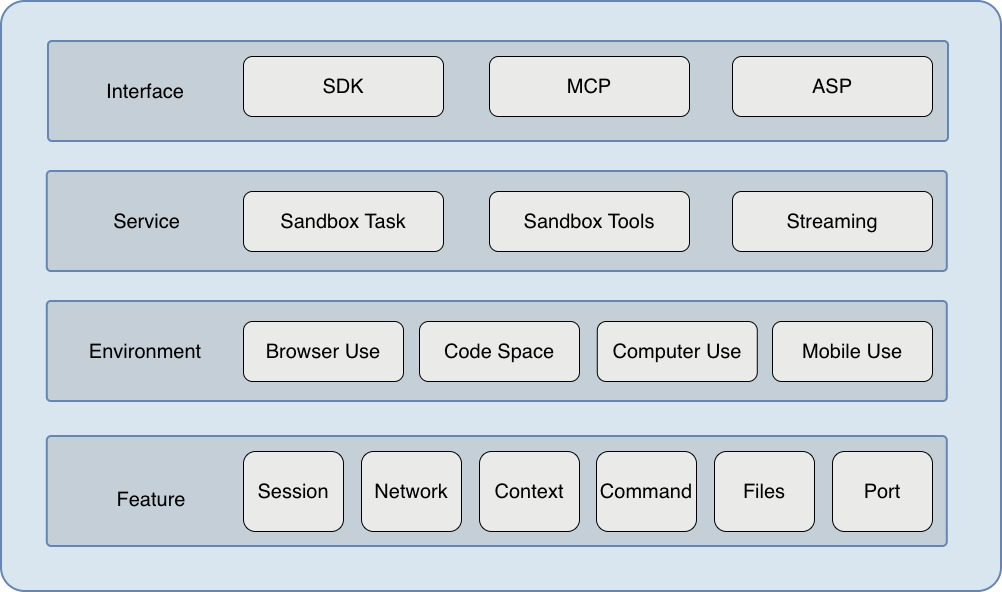}
  \caption{System architecture of AgentBay.}
  \label{fig:system-architecture}
\end{figure}

\subsection{Security}

Built upon a robust and highly stable infrastructure, the AgentBay service is architected with a security-first approach. Security is the fundamental pillar of the platform, centered on strict resource isolation where each agent session operates in a completely segregated environment. Given that AgentBay is designed to execute potentially untrusted, autonomously-generated code, this secure-by-design posture is paramount. Our security architecture is founded on the principles of zero trust and defense-in-depth, creating a holistic framework that addresses threats at both the infrastructure and agent level. This comprehensive strategy is realized through three core pillars: strong environmental isolation, a secure access plane that protects all data in transit, and a clearly defined shared responsibility model that delineates platform and user security duties.

\noindent\textbf{Strong Environmental Isolation:} AgentBay sandbox operates within its own dedicated virtual machine. This isolation is extended across storage, with each session possessing a private filesystem, and network, where sessions reside in isolated VPCs with a default-deny policy for inter-session and inbound communication. Critically, all sessions are ephemeral by design; upon termination, the entire virtual environment is destroyed, irretrievably wiping all transient data and state.

\noindent\textbf{Secure, Unified Access Plane:} All interactions with the sandbox, whether programmatic or manual, are funneled through a hardened gateway that serves as the sole, trusted entry point. For interactive human supervision, all graphical streams are transmitted via our Adaptive Streaming Protocol (ASP), which is tunneled through the secure gateway and benefits from end-to-end TLS encryption. This ensures that the high-performance stream for human control inherits the same rigorous security protections as programmatic API calls. The gateway further isolates the internal execution environment, mitigates common web-based attacks via an integrated WAF, and relies on short-lived, permission-scoped tokens to enforce the principle of least privilege for all access.

\noindent\textbf{Shared Responsibility and Agent-Level Security:} AgentBay is fundamentally responsible for the security of the sandbox, guaranteeing the integrity and isolation of the runtime environment itself. Conversely, the customer is responsible for the security in the sandbox and of the external agent that orchestrates the tasks. This customer-side responsibility includes securing the agent's core logic, its behavioral patterns, and its resilience against application-layer attacks such as prompt injection. This clear delineation, combined with the hardened environment and secure access protocols, creates a comprehensive security posture necessary for the safe deployment of autonomous agents.

\begin{figure}[htbp]
  \centering
  \includegraphics[width=0.9\linewidth]{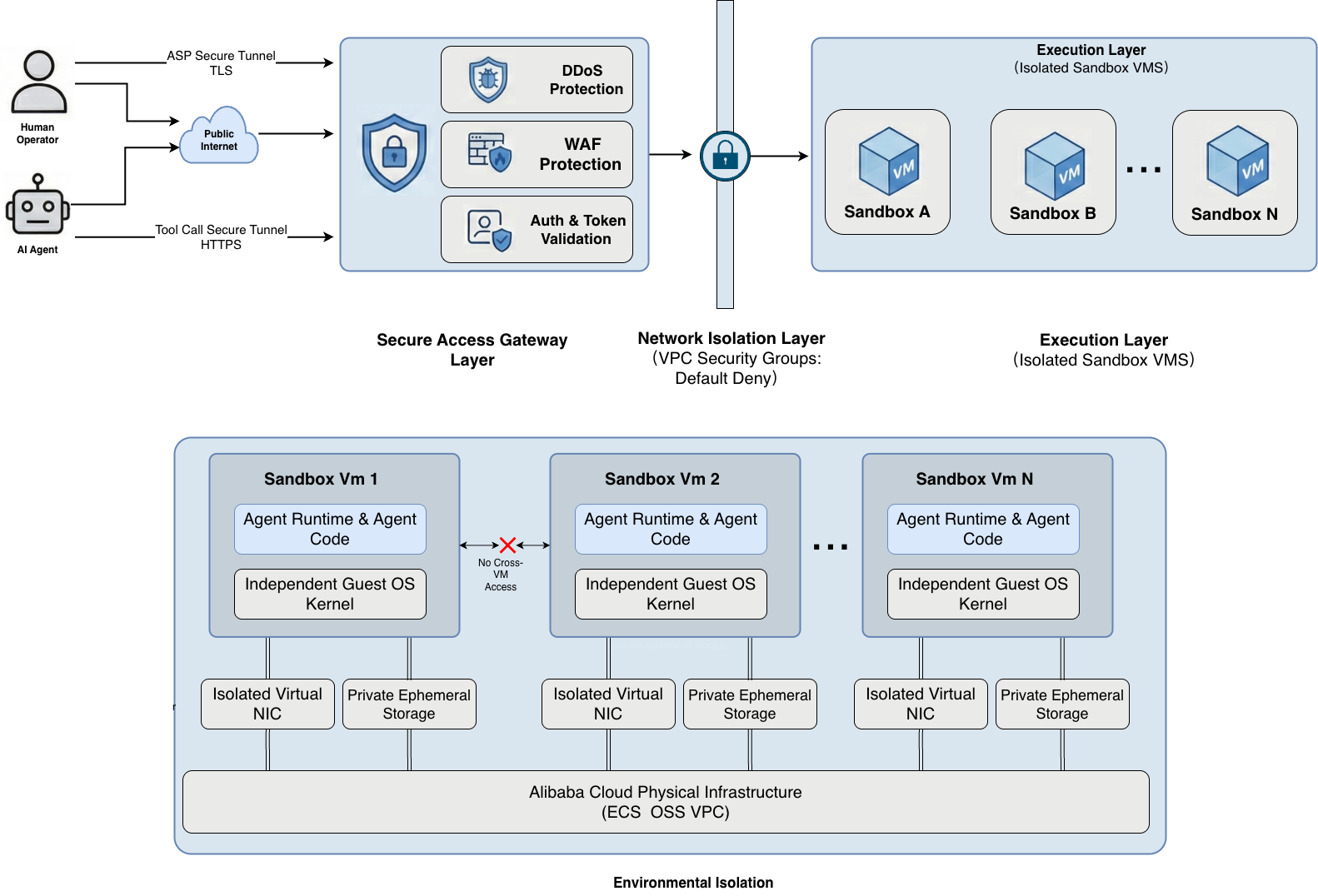}
  \caption{AgentBay Security Design.}
  \label{fig:security-design}
\end{figure}

\subsection{Adaptive Streaming Protocol (ASP)}

ASP centers on adaptive streaming to balance ultra-low latency with exceptional picture quality, delivering a stable and smooth client--cloud integrated real-time interactive experience.

\begin{figure}[htbp]
  \centering
  \includegraphics[width=0.9\linewidth]{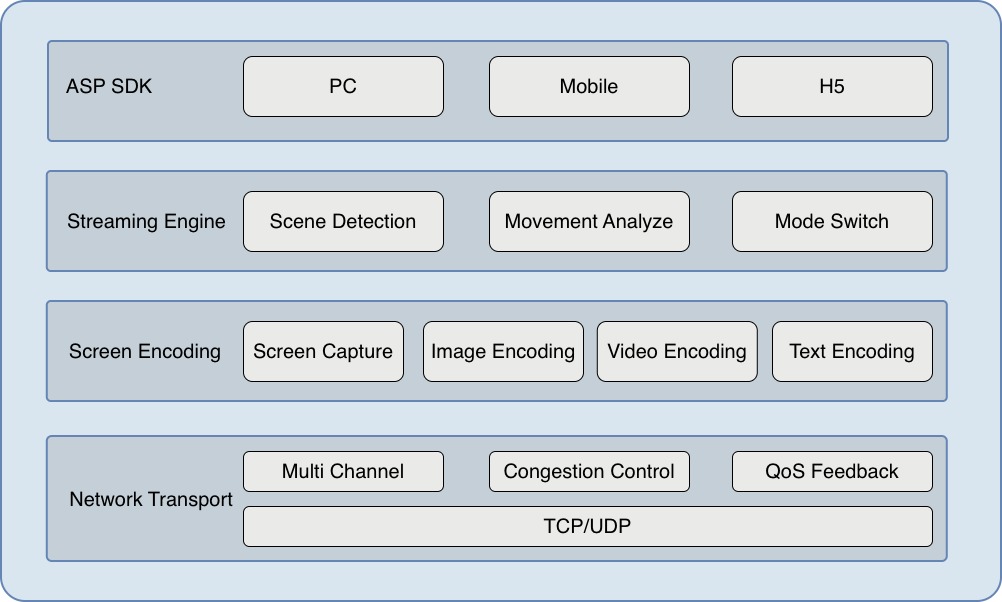}
  \caption{Adaptive Streaming Protocol.}
  \label{fig:asp}
\end{figure}

ASP synthesizes key technologies including visual content analysis, adaptive compression, real-time streaming, and network QoS optimization to deliver the following core features.

\begin{itemize}
  \item \textbf{Intelligent Streaming Engine:} The streaming pipeline encompasses screen capture, encoding, network transmission, and client-side rendering. To handle diverse workloads, the ASP engine distinguishes between low-motion scenarios (e.g., text editing, web browsing) and high-motion scenarios (e.g., video playback, gaming). It intelligently selects the optimal streaming path for each context, supporting both standard and GPU-accelerated cloud desktop environments.

  \item \textbf{Advanced Adaptive Compression:} Cloud sandbox screens typically display hybrid content, mixing computer-generated imagery (e.g., text, UI elements) with natural imagery (e.g., photos, videos). Relying solely on standard video or image codecs fails to balance quality and bandwidth for such mixed media. ASP addresses this by employing region-aware adaptive encoding. By analyzing screen updates, it classifies content types and transmits only dirty (changed) regions using the most appropriate algorithm. This approach significantly reduces bandwidth consumption while maintaining high visual fidelity.

  \item \textbf{Robust Network Transport:} Network jitter and packet loss often degrade real-time interaction. ASP mitigates these challenges through a multi-layered optimization strategy:
  \begin{itemize}
    \item Virtual multi-channel multiplexing prioritizes critical data packets over a single connection.
    \item A hybrid transport protocol leverages both TCP and UDP to establish a robust QoS foundation, incorporating dynamic bandwidth estimation and congestion control tailored to current network conditions.
    \item Audio optimization implements advanced buffering and jitter management on both uplink and downlink paths to ensure audio clarity.
  \end{itemize}

  \item \textbf{Unified Multi-platform SDK:} The ASP SDK is built upon a unified core library handling packet parsing, session management, input event upstreaming, and A/V decoding. Its cross-platform architecture supports a wide range of endpoints, enabling seamless, anytime access to ASP-enabled cloud services.
\end{itemize}

\section{Evaluation}

We conducted a comprehensive series of experiments to evaluate the AgentBay service, focusing on security robustness, HITL task execution, standard agent benchmarks, and ASP performance.

\subsection{Security Test}

To rigorously validate the isolation capabilities of AgentBay, we conducted a comparative security analysis against a standard local execution environment (Native Baseline). The evaluation focuses on two critical threat models inherent to autonomous agent deployment: host system integrity compromise and unauthorized network exfiltration.

We established two distinct execution environments for this evaluation.

\begin{itemize}
  \item \textbf{Native Baseline:} A standard Linux development environment with user-level privileges, simulating a typical local setup where agents operate directly on the host OS.
  \item \textbf{AgentBay Sandbox:} An isolated session instantiated within the AgentBay infrastructure, governed by our VM architecture and default-deny network policies.
\end{itemize}

We designed two adversarial attack vectors to simulate malicious or hallucinated agent behaviors:

\begin{itemize}
  \item \textbf{Vector A (Destructive Operations):} The agent injects the recursive deletion command ``rm -fr /'' to target the root directory and attempts to access host-level environment variables.
  \item \textbf{Vector B (Data Exfiltration):} The agent visits a compromised URL containing a script designed to harvest local tokens and exfiltrate them to an unauthorized external Command and Control server via \texttt{curl}.
\end{itemize}

Across both threat vectors, AgentBay consistently demonstrated stronger isolation and network protection compared to the Native Baseline.

\begin{itemize}
  \item \textbf{System Integrity and Isolation:} In the Native Baseline, Vector A caused irreversible system instability and exposed host credentials. Conversely, the AgentBay Sandbox strictly confined the destructive operations within an isolated filesystem. The environment-level isolation ensured that the underlying host infrastructure and concurrent tenant sessions remained completely unaffected. Furthermore, environment sanitization prevented the leakage of any host-level secrets, verifying the effectiveness of our process and filesystem isolation.

  \item \textbf{Network Governance:} Regarding Vector B, the Native Baseline failed to prevent outbound connections, resulting in confirmed data exfiltration. AgentBay, however, successfully intercepted the unauthorized traffic. Its default-deny network policy blocked the connection to the unverified C2 IP address. This effectively contained the sensitive data within the secure execution boundary, validating the robustness of the egress filtering mechanism.
\end{itemize}

\subsection{HITL Task Test}

We conducted web automation tests using a ReAct agent built with the Claude Sonnet 4.5 model in Chrome browser, evaluating the effectiveness of human takeover mechanisms across three typical automation failure scenarios.

\noindent\textbf{Testing Environment:}

\begin{itemize}
  \item \textbf{Agent Model:} Claude Sonnet 4.5
  \item \textbf{Automation Framework:} ReAct Agent + Chrome Browser
  \item \textbf{Test Scenarios:} Floating ad obstruction, CAPTCHA handling, password input
\end{itemize}

\noindent\textbf{Comparison Modes:}

\begin{itemize}
  \item \textbf{Agent-Only:} Fully autonomous execution with no human intervention.
  \item \textbf{Hybrid Interaction:} Agent can request human takeover when encountering difficulties.
\end{itemize}

\begin{table}[htbp]
  \centering
  \caption{Task success rates of Claude Sonnet 4.5 agent across three scenarios.}
  \label{tab:hitl-task}
  \begin{tabular}{@{}lccc@{}}
    \toprule
    Scenario & Agent-Only Success Rate & Hybrid Interaction Success Rate & Improvement \\ \midrule
    Text reading under floating ad obstruction & 27\% & 97\% & +259\% \\
    Web CAPTCHA handling                       & 64\% & 95\% & +48\%  \\
    Password input                             & --  & 100\% & --     \\ \bottomrule
  \end{tabular}
\end{table}

We analyze the specific failure patterns observed in the HITL task test.

\begin{enumerate}
  \item \textbf{Floating Ad Obstruction (Highest Failure Rate: 73\%)}\\
  Dynamic advertisements cover target text or buttons, preventing the agent from accurately locating interactive elements. After human takeover, ad pop-ups can be quickly closed (average 5--8 seconds), restoring normal page state. This aligns with industry observations: modern websites deploy anti-ad-blocker mechanisms that cause key UI elements to fail loading when automation tools are detected.

  \item \textbf{Web CAPTCHA Handling (Medium Failure Rate: 36\%)}\\
  Despite the model having visual understanding capabilities, 36\% of CAPTCHA recognition attempts still fail. Human takeover enables direct verification completion (average 15--20 seconds). Research data supports this finding: general AI agents face 60\% failure rates on modern CAPTCHA systems, and even our LLM-enhanced specialized solvers only achieve 60+\% success rates in mixed CAPTCHA scenarios~\cite{placeholder-captcha-benchmark}.

  \item \textbf{Password Input (Absolute Failure: 100\%)}\\
  For security reasons, the agent cannot access user password credentials. After users input passwords directly through the sandbox, the agent continues subsequent workflows. This is an inherent result of architectural design, reflecting the principle of reserving sensitive operations for humans.
\end{enumerate}

\subsubsection*{Benchmark Comparison Data}

Our results are consistent with the latest agent benchmark tests.

\noindent\textbf{Web Automation Benchmarks (2025 Latest Data):}

As of February 2025, the highest single-agent task completion rate on the WebArena benchmark reached 61.7\% (IBM CUGA~\cite{placeholder-webarena-cuga}), while human performance remains at 78\%. Over two years, AI agent success rates on WebArena leaped from 14\% to approximately 60\%, but still show a significant gap from human-level performance (78\%).

On the more challenging WebChoreArena benchmark, despite top LLMs (such as Gemini 2.5 Pro) reaching 54.8\% on the original WebArena, they only achieve 37.8\% on WebChoreArena, demonstrating that complex, long-term memory tasks remain a major challenge for autonomous agents~\cite{placeholder-webchorearena}.

The experiments demonstrate that even using the most advanced multimodal model (Claude Sonnet 4.5), purely autonomous agents still have significant limitations in real-world web automation scenarios:

\begin{enumerate}
  \item \textbf{Dynamic UI elements} (floating ads) cause 73\% reading failures.
  \item \textbf{Adversarial mechanisms} (CAPTCHAs) result in 36\% handling failures.
  \item \textbf{Security boundaries} (password input) require 100\% human participation.
\end{enumerate}

The hybrid interaction mode implemented through sandbox takeover mechanisms improved success rates to over 95\% across all three scenarios, with average human intervention times of only 15--30 seconds. This validates that selective human intervention is a key strategy for improving agent reliability, not a stopgap measure. Even in the most advanced benchmark tests of 2025, purely autonomous agents in complex real-world scenarios still achieve far lower success rates than the 95\%+ success rates that human-in-the-loop systems achieve in production environments.

\subsection{Open-Source Agent Benchmark}

SeeAct~\cite{placeholder-seeact} is a generalist web agent system designed to autonomously perform tasks on any website using large multimodal models (LMMs) such as GPT-4V. The system consists of two main components: a robust codebase for running web agents on live websites, and an innovative framework that utilizes LMMs as generalist web agents. It serves as an interface between an agent and a web browser, efficiently transmitting inputs from the browser to the agent and translating the agent's predicted actions into browser events for execution.

Online-Mind2Web~\cite{placeholder-online-mind2web} is a benchmark platform for evaluating web agents in real online environments, extending the original Mind2Web dataset to address the ``progress illusion'' problem in current research. It comprises 300 diverse tasks across 136 popular websites spanning domains like fashion, food, housing, and transportation. Designed to overcome key evaluation challenges---such as low alignment between automated metrics and human judgment, and task failures due to dynamic web changes (e.g., site redesigns or CAPTCHAs)---it introduces WebJudge, an LLM-based three-stage evaluation framework. WebJudge first identifies critical task checkpoints from instructions, then selects relevant screenshots from agent trajectories to preserve visual evidence while reducing noise, and finally judges task success by integrating task descriptions, key points, key screenshots, and action history---ensuring reliable, scalable, and token-efficient assessment.

To validate hybrid interaction as a general-purpose agent testbed, we ran SeeAct---a popular open-source agent implementation---on the public Online-Mind2Web benchmark, aiming to demonstrate that our environment faithfully reproduces results and serves as a reliable testbed. For rapid consistency verification, we selected 14 accessible tasks of the ``Easy'' difficulty from Online-Mind2Web, along with their corresponding websites, and executed them on both a physical machine and the remote hybrid interaction environment.

Based on the test results below, tasks run in the AgentBay sandbox achieve higher success scores compared to those run on physical machines, proving that it is more user-friendly for agents during use.

\begin{table}[htbp]
  \centering
  \setlength{\tabcolsep}{4pt}
  \caption{SeeAct agent on Online-Mind2Web benchmark.}
  \label{tab:open-source-benchmark}
  \begin{tabular}{@{}lcccc@{}}
    \toprule
    Benchmark & Model & Browser Agent & Score in Physical Machine & Score in AgentBay \\ \midrule
    OnlineMind2Web (Easy, 14 Tasks) & qwen3-vl-plus & SeeAct & 35.71\% & 42.86\% \\ \bottomrule
  \end{tabular}
\end{table}

\subsection{ASP Performance}

We compared ASP against RDP in four key areas: interactive latency, stutter rate, bandwidth consumption, and picture quality.

\subsubsection*{Interactive Latency}

We measured ``click-to-photon'' latency (time from mouse click to seeing the result on-screen).

\noindent\textbf{Results:} ASP's hybrid-graphic-based mode achieves latency comparable to RDP, and is around 5\% faster than RDP.

\begin{table}[htbp]
  \centering
  \setlength{\tabcolsep}{20pt}
  \caption{Latency comparison of ASP and RDP.}
  \label{tab:latency}
  \begin{tabular}{@{}lc@{}}
    \toprule
    Protocol & Average latency \\ \midrule
    ASP      & 117ms \\
    RDP      & 122ms \\ \bottomrule
  \end{tabular}
\end{table}

\subsubsection*{Stutter Rate}

We measured the proportion of late frames (frame time exceeding the target threshold) over 60 seconds across common tasks (browsing web pages and video playback) with different network qualities.

\noindent\textbf{Results:} Across nearly all scenarios, ASP achieved a much lower stutter rate than RDP.

\begin{table}[htbp]
  \centering
  \setlength{\tabcolsep}{8pt}
  \caption{Stutter rate comparison of ASP and RDP.}
  \label{tab:stutter}
  \begin{tabular}{@{}p{0.32\textwidth} p{0.18\textwidth} p{0.10\textwidth} p{0.12\textwidth} p{0.12\textwidth}@{}}
    \toprule
    Network Quality & Scenario & FPS & ASP & RDP \\ \midrule
    \multirow{2}{*}{Normal} & Browse web page & 20fps & 0.51\%  & 1.16\%  \\
    & Video playback  & 30fps & 0.59\%  & 5.08\%  \\
    \multirow{2}{*}{Bandwidth restricted to 5Mbps} & Browse web page & 20fps & 1.29\%  & 36.45\% \\
    & Video playback  & 30fps & 2.57\%  & 7.57\%  \\
    \multirow{2}{*}{Downstream packet drop rate 10\%} & Browse web page & 20fps & 16.45\% & 96.16\% \\
    & Video playback  & 30fps & 8.55\%  & 44.96\% \\ \bottomrule
  \end{tabular}
\end{table}

\subsubsection*{Bandwidth Consumption}

We measured average bandwidth over 60 seconds for four common tasks.

\noindent\textbf{Results:} ASP adapts to be the most efficient in all scenarios. It behaves like RDP for static text in Word scenarios, but achieves a substantial lead over RDP in all other scenarios.

\begin{table}[htbp]
  \centering
  \setlength{\tabcolsep}{16pt}
  \caption{Bandwidth consumption comparison of ASP and RDP.}
  \label{tab:bandwidth}
  \begin{tabular}{@{}lcc@{}}
    \toprule
    Scenario        & ASP     & RDP      \\ \midrule
    Use Word        & 85kbps  & 99kbps   \\
    Use PowerPoint  & 1.4mbps & 2mbps    \\
    Browse web page & 1.3mbps & 1.9mbps  \\
    Play video      & 4.6mbps & 10.2mbps \\ \bottomrule
  \end{tabular}
\end{table}

\subsubsection*{Picture Quality}

We measured picture quality for four common tasks.

\noindent\textbf{Results:} Aside from static text operations (e.g., Word), where RDP has a slight edge, ASP outperforms RDP in all other scenarios.

\begin{table}[htbp]
  \centering
  \setlength{\tabcolsep}{16pt}
  \caption{Picture quality comparison of ASP and RDP.}
  \label{tab:quality}
  \begin{tabular}{@{}lcc@{}}
    \toprule
    Scenario        & ASP   & RDP   \\ \midrule
    Use Word        & 0.562 & 0.569 \\
    Browse web page & 0.833 & 0.827 \\
    View image      & 0.960 & 0.959 \\
    Play video      & 0.753 & 0.750 \\ \bottomrule
  \end{tabular}
\end{table}

\section{Conclusion}

We presented AgentBay, a hybrid interaction sandbox system that resolves the false dichotomy between programmatic control and human interaction. By providing a single, unified session accessible via both a programmatic API and an ultra-low-latency graphical stream, AgentBay enables seamless human supervision of autonomous agents. Our core contribution, the Adaptive Streaming Protocol (ASP), demonstrates superior performance by dynamically adapting its streaming strategy based on the current controller.

Our evaluation demonstrates strong results across multiple dimensions. In security testing, AgentBay's multi-tenant architecture with strict isolation successfully prevents cross-session interference and unauthorized access, effectively containing destructive operations and blocking unauthorized network exfiltration attempts. In task completion experiments, our evaluation across multiple benchmarks reveals the substantial impact of human-in-the-loop supervision: the hybrid interaction model (Agent + Human) improves success rates more than 48\%, compared to agent-only systems. This dramatic improvement validates the necessity of human-in-the-loop supervision for reliable autonomous agent deployment. Furthermore, our ASP protocol demonstrates superior performance: it achieves interactive latency of 117ms (approximately 5\% faster than RDP's 122ms), reduces bandwidth consumption by up to 55\% compared to RDP in video playback scenarios (4.6mbps vs 10.2mbps), while maintaining comparable or superior picture quality. These results demonstrate that specialized protocols can achieve enhanced efficiency when tailored to specific application scenarios.

Our results strongly indicate that a dual-modal infrastructure is essential for deploying reliable agents. The ``all-or-nothing'' approach to autonomy is too brittle for real-world applications. By providing a ``human escape hatch'' that is seamless, we lower the cost of failure and increase overall system robustness. The evaluation results demonstrate that agent failures manifest systematic and predictable patterns, including modal dialogs, CAPTCHAs, dynamic content, and ambiguous visual states. These failure modes present fundamental challenges for programmatically-driven agents operating autonomously, as they necessitate contextual comprehension, visual reasoning mechanisms, or explicit human judgment. The seamless handoff mechanism implemented in our system addresses this limitation by elevating human intervention to a first-class architectural component rather than treating it as an exceptional case, thereby transforming HITL from an ad-hoc solution into a fundamental architectural feature.

For mission-critical applications---including financial transactions, healthcare data management, and enterprise software deployment---pure autonomy may prove inadequate. The hybrid interaction model proposed in this work provides a viable alternative: agents maintain autonomous operation for routine tasks, while human operators can intervene seamlessly when exceptional circumstances arise or when sensitive operations necessitate explicit authorization. The security and isolation guarantees provided by AgentBay are critical prerequisites for enterprise deployment. Organizations deploying autonomous agents must reconcile the benefits of automation against the risks of unauthorized access or data leakage. Our multi-tenant architecture, incorporating strict process, filesystem, and network isolation, mitigates these security concerns, enabling the deployment of agents in production environments where security requirements are stringent.

In conclusion, AgentBay addresses a fundamental gap in the autonomous agent ecosystem by providing a secure, high-fidelity, and seamlessly controllable execution environment. The hybrid interaction model we propose transforms human intervention from an exception into a core architectural feature, thereby enabling the deployment of reliable, human-supervised autonomous systems. We posit that AgentBay provides a foundational primitive for building the next generation of reliable, autonomous systems that can operate effectively in real-world scenarios where brittleness and exceptions are inevitable.


\renewcommand{\refname}{References}


\begin{thebibliography}{99}

\bibitem{placeholder-llm-agent-survey}
Z.~Xi et al., ``A Survey on Large Language Model based Autonomous Agents,'' \emph{arXiv preprint} arXiv:2308.11432, 2023.

\bibitem{placeholder-autogpt}
Significant Gravitas, ``Auto-GPT: An Autonomous GPT-4 Experiment,'' GitHub repository, 2023. Available at: \url{https://github.com/Significant-Gravitas/Auto-GPT}.

\bibitem{placeholder-langchain}
H.~Chase, ``LangChain,'' GitHub repository, 2022. Available at: \url{https://github.com/langchain-ai/langchain}.

\bibitem{placeholder-llamaindex}
J.~Liu, ``LlamaIndex,'' \emph{GitHub repository}, 2022. [Online]. Available: https://github.com/run-llama/llama\_index

\bibitem{placeholder-webarena}
T.~Zhou, N.~Zhang, and B.~Zong, ``WebArena: A Realistic Web Environment for Building and Testing Autonomous Agents,'' in \emph{ICLR}, 2024.

\bibitem{placeholder-hitl-survey}
T.~Wu, L.~Aroyo, and C.~Welty, ``A Survey on Human-in-the-Loop for Machine Learning,'' \emph{arXiv preprint} arXiv:1904.03723, 2019.

\bibitem{placeholder-e2b}
E2B, ``E2B: Secure Sandboxes for AI Agents,'' 2023. Available at: \url{https://e2b.dev}.

\bibitem{placeholder-vnc}
T.~Richardson, Q.~Stafford-Fraser, K.~R.~Wood, and A.~Hopper, ``Virtual Network Computing,'' \emph{IEEE Internet Computing}, vol.~2, no.~1, pp.~33--38, 1998.

\bibitem{placeholder-rdp}
Microsoft Corporation, ``Remote Desktop Protocol (RDP),'' Microsoft Documentation, 2020.

\bibitem{placeholder-agentbay-sdk}
Alibaba~Cloud~Computing, ``Wuying AgentBay SDK,'' \emph{GitHub repository}, 2025. [Online]. Available: https://github.com/aliyun/wuying-agentbay-sdk

\bibitem{placeholder-webrtc}
A.~B.~Johnston and D.~C.~Burnett, ``WebRTC: APIs and RTCWEB Protocols of the HTML5 Real-Time Web,'' Digital Codex, 2012.

\bibitem{placeholder-agentbench}
S.~Liu, H.~Ye, and K.~Ji, ``AgentBench: A Comprehensive Benchmark for LLM-as-Agent,'' \emph{arXiv preprint} arXiv:2308.03688, 2023.

\bibitem{placeholder-react}
S.~Yao et al., ``ReAct: Synergizing Reasoning and Acting in Language Models,'' in \emph{ICLR}, 2023.

\bibitem{placeholder-daytona}
Daytona, ``Daytona: Secure, standard, and automated dev environments,'' 2023. Available at: \url{https://daytona.io}.

\bibitem{placeholder-kvm}
A.~Kivity et al., ``KVM: the Linux virtual machine monitor,'' in \emph{Proceedings of the Linux Symposium}, vol.~1, pp.~225--230, 2007.

\bibitem{placeholder-codeagent}
X.~Tang et al., ``CodeAgent: Autonomous Communicative Agents for Code Review,'' \emph{arXiv preprint} arXiv:2402.02172, 2024.

\bibitem{placeholder-ds-agent}
P.~Wang et al., ``Large Language Model-based Data Science Agent: A Survey,'' University of Illinois Urbana-Champaign, Tech. Rep., 2025, \emph{arXiv}:2508.02744.

\bibitem{placeholder-auto-arena}
R.~Zhao et al., ``Auto-Arena: Automating LLM Evaluations with Agent Peer Battles and Committee Discussions,'' \emph{arXiv preprint} arXiv:2405.20267, 2024.

\bibitem{placeholder-phone-ui}
N.~Kahlon et al., ``Agent-Initiated Interaction in Phone UI Automation,'' \emph{arXiv preprint} arXiv:2503.19537, 2025.

\bibitem{placeholder-game-ai}
C.~Hu et al., ``Games for Artificial Intelligence Research: A Review and Perspectives,'' \emph{arXiv preprint} arXiv:2304.13269, 2024.

\bibitem{placeholder-captcha}
S.~Anupam et al., ``BrowserArena: Evaluating LLM Agents on Real-World Web Navigation Tasks,'' \emph{arXiv preprint} arXiv:2510.02418, 2025.

\bibitem{placeholder-captcha-benchmark}
T.~Zhang, Z.~Wang, H.~Duan, and J.~Crk, ``Oedipus: LLM-enhanced Reasoning CAPTCHA Solver,'' in \emph{Proceedings of the 2025 ACM SIGSAC Conference on Computer and Communications Security (CCS)}, 2025.

\bibitem{placeholder-webarena-cuga}
S.~Marreed et al., ``Towards Enterprise-Ready Computer Using Generalist Agent,'' \emph{arXiv preprint} arXiv:2503.01861, 2025.

\bibitem{placeholder-webchorearena}
A.~Miyai et al., ``WebChoreArena: Evaluating Web Browsing Agents on Realistic Tedious Web Tasks,'' \emph{arXiv preprint} arXiv:2506.01952, 2025.

\bibitem{placeholder-seeact}
B.~Gou et al., ``SeeAct: GPT-4V(ision) is a Generalist Web Agent, if Grounded,'' in \emph{Proceedings of the 2024 Conference on Empirical Methods in Natural Language Processing (EMNLP)}, 2024.

\bibitem{placeholder-online-mind2web}
X.~Deng et al., ``On the Progress Illusion of Web Agents: The Online-Mind2Web Benchmark,'' \emph{arXiv preprint} arXiv:2410.15049, 2024.

\end{thebibliography}
\end{document}